\pdfoutput=1

\documentclass[11pt]{article}

\usepackage[]{acl}

\usepackage{times}
\usepackage{latexsym}
\usepackage{tabularx}

\usepackage[T1]{fontenc}

\usepackage[utf8]{inputenc}

\usepackage{microtype}

\usepackage{inconsolata}

\usepackage{amsmath}
\usepackage{graphicx}
\usepackage{booktabs}
\usepackage{cleveref}
\usepackage{color, colortbl}
\usepackage{multirow}
\usepackage{algorithm}
\usepackage[noend]{algpseudocode}

\usepackage{array}

\usepackage{xcolor}
\definecolor{mypink1}{rgb}{0.858, 0.188, 0.478}
\definecolor{darkred}{rgb}{0.74,0.03,0}
\definecolor{mustardyellow}{rgb}{0.88,0.67,0.01}
\definecolor{navy}{rgb}{0,0,0.5}
\definecolor{darkcyan}{rgb}{0,0.54,0.54}
\definecolor{tabhighlight}{rgb}{0,0.54,0.54}

\title{Latent Factor Models Meets Instructions:\\Goal-conditioned Latent Factor Discovery without Task Supervision}

\def\authorspace{\hspace{4mm}}
\def\ucimark{$^1$}
\def\uomark{$^2$}

\author{
    Zhouhang Xie\ucimark{}\authorspace{}
    Tushar Khot\uomark{}{}\authorspace{}
    Bhavana Dalvi Mishra\uomark{}\authorspace{}
    Harshit Surana\uomark{}\authorspace{}
    \\
    \textbf{
    Julian McAuley\ucimark{}\authorspace{}
    Peter Clark\uomark{}\authorspace{}
    Bodhisattwa Prasad Majumder\uomark{}\authorspace{}
    }
        \\
        \ucimark{}~University of California, San Diego \\
        \uomark{}~Allen Institute for AI \\
        \href{mailto:zhx022@ucsd.edu}{\texttt{zhx022@ucsd.edu}}, \href{mailto:bodhisattwam@allenai.org}{\texttt{bodhisattwam@allenai.org}} \\
        Website: \url{https://github.com/allenai/instructLF}
        }

\newcommand{\ourmethod}{Instruct-LF}

\begin{document}
\maketitle

\begin{abstract}

Instruction-following LLMs have recently allowed systems to discover hidden concepts from a collection of unstructured documents based on a natural language description of the purpose of the discovery (i.e., goal). 
Still, the quality of the discovered concepts remains mixed, as it depends heavily on LLM's reasoning ability and drops when the data is noisy or beyond LLM's knowledge.
We present \ourmethod, a goal-oriented latent factor discovery system that integrates LLM's instruction-following ability with statistical models to handle large, noisy datasets where LLM reasoning alone falls short.

\ourmethod~uses~LLMs to propose fine-grained, goal-related properties from documents, estimates their presence across the dataset, and applies gradient-based optimization to uncover hidden factors, where each factor is represented by a cluster of co-occurring properties. 
We evaluate latent factors produced by \ourmethod~on movie recommendation, text-world navigation, and legal document categorization tasks. 
These interpretable representations improve downstream task performance by 5-52\% than the best baselines and were preferred 1.8 times as often as the best alternative, on average, in human evaluation.

\end{abstract}

\section{Introduction}

Algorithms for discovering interpretable latent structures from observed data is a long-standing challenge in AI~\cite{Fayyad1996TheKP}, with applications to bioinformatics~\cite{liu2016overviewtopicbio}, social science~\cite{ramage2009topic}, e-commerce~\cite{mcauley2013hidden}, and beyond.
These methods help users make sense of large amounts of unstructured data to draw insights for various needs.

\begin{figure}[tb]
\centering
\includegraphics[scale=0.65]{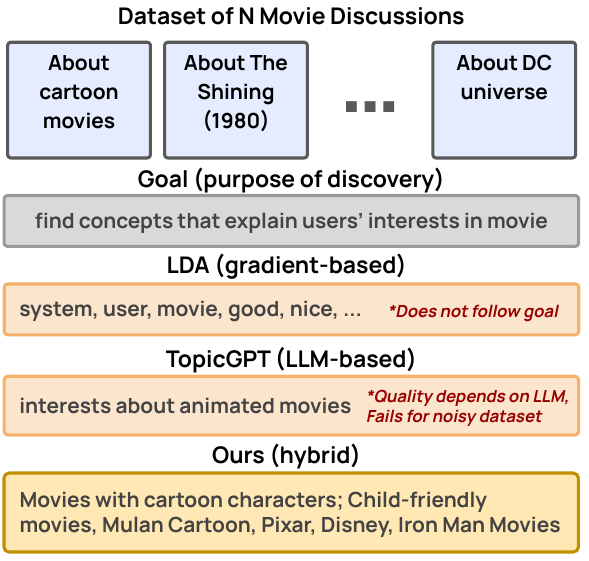}
\caption{
An example of the output of our system,~\ourmethod, on discovering different types of user interest about movies from a conversation corpus. 
}
\label{fig:motivation}
\vskip -5mm
\end{figure}

We seek to develop an instruction-following latent factor discovery system that can adapt its discovery process based on user instructions.
For example, as shown in~\Cref{fig:motivation}, a movie streaming platform (the user) may wish to analyze a dialogue corpus where its customers chat about movies, with the goal of gaining insights into different types of movies their customers are interested in.
In this case, the latent factor discovery system should ideally discard properties in the data that are irrelevant to the platform's goal, for example, article words in LDA~\cite{Blei2009LatentDA}'s output.
Meanwhile, the system should also produce informative latent factors with fine-grained details, which existing LLM-based frameworks (e.g., \citet{pham-etal-2024-topicgpt}) cannot consistently generate for noisy data.

Importantly, we focus on the case where we do not rely on the properties of specific datasets to guide gradient-based latent factor models.  
Such a constraint makes setting up task-specific supervision signals, e.g., auxiliary loss functions that exploit the structure of the corpus (such as sentiment labels or co-clicks associated with each document~(\citet{mcauley2013hidden, mcauley2015inferring}) infeasible.%

Classic latent factor models such as Latent Dirichlet Allocation (LDA)~\cite{Blei2009LatentDA} and BertTopic~\cite{ Grootendorst2022BERTopicNT} are popular choices in mining latent patterns from data, however, they cannot flexibly follow user instructions.
Further, interpreting the latent space of classical latent factor models frequently requires reading-the-tea-leaves interpretations, such as examining clusters of keywords~\cite{Blei2009LatentDA} or interpreting sampled documents~\cite{Kingma2013AutoEncodingVB}.
As we will later show in our experiments, these noisy explanations only provide good interpretability for users when the most salient signals (i.e., words) in the dataset align with the user goal well.

Another recent emerging paradigm is to prompt LLMs for pattern discovery from documents~\cite{wang-etal-2023-goal, Zhong2023D5, pham-etal-2024-topicgpt}. 
While these LLM-based frameworks can adapt their behavior based on user instructions, these methods do not scale well to large-scale, noisy datasets.
In particular, LLM-based methods generally first prompt LLMs to generate potentially interesting pattern descriptions, such as topic names, prompted with the input data, then use LLMs to link data points to these generated descriptions~\cite{wang-etal-2023-goal, pham-etal-2024-topicgpt}.
However, since such a process is purely LLM-driven, its success is conditioned on an LLM's ability to reason over the dataset of interest, which fails when the observed data exceeds the content understanding ability of LLMs~\cite{pham-etal-2024-topicgpt}.
As we will later show in our experiments, when dealing with noisy, out-of-distribution data, these pure LLM-based methods fail to consistently produce coherent results that are helpful for users.

To address these challenges, we propose~\textbf{\ourmethod}\footnote{\url{https://github.com/allenai/instructLF}}, a framework that combines the \textbf{Instruct}ion-following task-adaptability of LLMs with classical gradient-based \textbf{L}atent \textbf{F}actor modeling algorithms.
Unlike existing works on using LLMs for topic proposal and assignment~\cite{pham-etal-2024-topicgpt, wang-etal-2023-goal}, we propose simplifying LLM-based operations in our framework to a minimum: an \texttt{property proposal} step that simply prompts LLMs for goal-related properties based on an input document, where each property is a natural language statement that describes a goal-oriented attribute of the input data point. 
After this, we estimate the occurrence of each candidate property across the dataset by factorizing and filling a sparsely filled data-property matrix, where an observed data-to-property linkage with high estimated score means a property is more likely generated from a data point via the \texttt{property proposal} step.
Finally, our framework clusters these properties into groups by estimating their correlation through their estimated occurrence across the dataset. 
The grouped properties then become explanations of the discovered latent factors, similar to how topic models use grouped keywords as discovered topics.

We evaluate our proposed method on three scenarios: (1) analyzing a dialogue corpus where users recommend movies to each other, with the goal of understanding different types of user interest, (2) analyzing user-environment interaction logs on a text-based world simulator, Alfworld~\cite{shridhar2021alfworld}, with the goal of discovering different meaningful states that are relevant to task completion, and (3) analyzing a set of American bill summaries, with the goal of discovering categories of bills.
The first two scenarios require a model that can adapt to users' goal of discovery, while the third provides a testbed where classic latent factor models are known to perform well.
Automatic and human evaluation show~\ourmethod~can discover informative and task-relevant patterns from data, and rated as the best-picked model across various baselines in human evaluation.

\textbf{Our contributions are as follows}: We develop a new method,~\ourmethod, that uncovers latent patterns that are relevant to the users' goal of discovery expressed in natural language from unstructured data.
Second,~\ourmethod~is the first work, to our best knowledge, to combine LLM reasoning with classical gradient-based methods for latent pattern mining.
Finally, we perform comprehensive evaluations and show that our system uncovers goal-relevant, coherent, informative, and interpretable latent factors from data and is chosen most frequently as the best model against state-of-the-art baselines in human evaluation.

\section{Related Work}

\begin{figure*}[tb]
\centering
\includegraphics[scale=0.65]{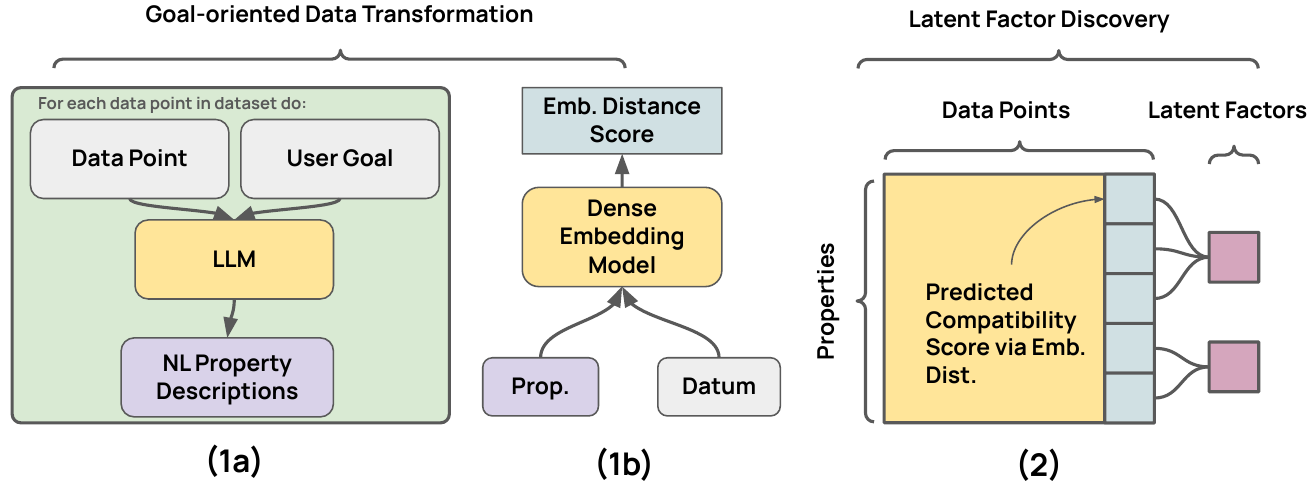}
\caption{The proposed framework.~\ourmethod~generates a set of natural language property descriptions from data, i.e., documents (1a); then estimates the compatibility between each data point and each property (1b), and perform correlation-based grouping of properties to discover latent factors (2). The compatibility between each property is efficiently computed using a distilled dense text representation model. We provide additional details and examples in~\Cref{fig:our_framework_detailed} and~\Cref{sec:our_framework_detailed}.
}
\label{fig:our_framework}
\vspace{-2mm}
\end{figure*}

\paragraph{Latent Factor Models}
Latent factor models (LFMs) assume observed data in a dataset are governed by a set of latent factors. 
Traditional algorithms for estimating latent factors from data such as LDA~\cite{Blei2009LatentDA}, PCA~\cite{Jolliffe2016PrincipalCA}, and Autoencoders~\cite{Rumelhart1986LearningIR}. 
Oftentimes, practitioners gain insight into the dataset by interpreting the learned factors, such as reading the topics in an LDA model or generating samples from the latent space of a variational autoencoder~\cite{Kingma2013AutoEncodingVB}.

\paragraph{LLMs for Insight Mining}
Recently, there have been some successes in using LLM-based systems for analyzing textual documents, such as using LLMs to discover topics from documents~\cite{pham-etal-2024-topicgpt} or clustering documents~\cite{wang-etal-2023-goal}.
The most relevant framework to our work is TopicGPT~\cite{pham-etal-2024-topicgpt}, which uses LLM to propose potential topics by sequentially iterating over the dataset and generating new topics based on prior outputs.

\paragraph{Key Differences} While drawing insprirations from the above areas,~\ourmethod~is the first work to combine LLMs' instruction-following ability with statistical models for goal-conditioned latent factor discovery.
In contrast to pure-LLM-based frameworks, our method only rely on LLMs to document interesting properties from data, which is less reliant on LLMs' reasoning ability.
By combining these LLM-proposed, goal-oriented properties with statistical-model-based latent factor models, we later show in our experiments that~\ourmethod~is the only method that can discover informative patterns from noisy data such as conversation and embodied navigation logs.
We provide further discussion on other related research areas in~\Cref{sec:further_related_work}.

\section{\ourmethod}

\paragraph{Overview} 
As shown in~\Cref{fig:our_framework},
there are two key steps in~\ourmethod. 
First, we perform a \textbf{goal-oriented data transformation} step where we generate a set of goal-related properties using LLMs and and transform the input unstructured dataset into a data-property matrix by learning a property-compatibility score for each document.
That is, each entry in this matrix is a compatibility score between a natural language property description and a data point, denoting how likely the property describes the data point.
Then, we perform a \textbf{latent factor discovery} step where we group correlated properties into clusters representing higher-level abstract concepts.

During goal-oriented data transformation, we learn a mapping function $g$ that maps the high-dimensional input space to a property space $g(X) \to C$, where each dimension $c_i \in C$ represents a goal-related property.
For example, when the user's (e.g., a streaming platform's) goal is to discover different types of customer interest in movies, a goal-related property can be ``\texttt{Disneyland Movies}", which is a property related to ``the genres of movies that interests customers."

We then perform the latent factor discovery step by learning a transformation $f(C) \to Z$, where $Z$ is a low-dimensional latent space that preserves the covariance in $C$.
I.e., properties with high correlations should be grouped together, which reflects higher-level, more abstract meanings; similar to how a set of words reflects a topic in topic models~\cite{Blei2009LatentDA, Grootendorst2022BERTopicNT}.
In this way, each dimension in $Z$ is a latent high-level concept corresponding to a set of dimensions in $C$, and can thus be explained by its associated properties in the property space.
Compared to recent works that rely on LLMs to directly propose and refine higher-level concepts~\cite{pham-etal-2024-topicgpt}, our method relies less on LLMs' reasoning ability, and can benefit from gradient-based learning from larger datasets.

\subsection{Goal-oriented Data Transformation}

Goal-oriented data transformation takes as input a set of unstructured input data and the users' description of the goal for discovery in natural language.
It then returns a matrix where each row represents a data point, and each column is a property related to the user's goal.
To achieve this, we implement an \textit{property proposal} step, where we prompt LLMs to generate a set of candidate properties, and a \textit{data-property link prediction} step, where we predict compatibility between each property and each data point.

\paragraph{Property Proposal} 
For each data point in the dataset, we prompt an LLM to generate a set of properties that describe the goal-oriented properties of this data point (see~\Cref{sec:implementation_details} for details).
While prior work rely on LLMs to generate high-level concepts that are comprehensive and generalizable over a batch of data~\cite{pham-etal-2024-topicgpt}, our \texttt{property proposal} step only prompt LLMs to document detailed attributes of a single data point (see~\Cref{sec:discussion_on_prompts} for additional details).

This formulation allows our framework to capture fine-grained details of the dataset (since the LLM can focus on details from only one datapoint), and is less demanding for an LLM's intrinsic ability to propose high-quality properties that generalize across the dataset. 

\paragraph{Data-property Link Prediction} 
After we collected a pool of properties in the property proposal stage, the next step is to estimate the occurrence of each seen property across the whole dataset.
Previous work generally prompt LLM to determine whether a property is relevant given a data point.
However, this formulation requires $N * |C|$ LLM calls, where $N$ is the number of data points of interest, making the algorithm inefficient (further discussions in~\Cref{sec:discussion_on_efficiency}).
Further, in our ablation experiments~(\Cref{sec:human_eval_and_ablations}), we find LLMs' ability to determine compatibility between properties and data lags behind its ability to generate plausible properties.
To this end, we propose to train a dual-embedding model to estimate the values in the data-property matrix by leveraging higher-quality linkages generated in the proposal stage as supervision signals.

Specifically, we adapt the widely used neural matrix factorization setting~\cite{he2017neuralCF}, where we learn a neural encoder $\Phi$ to estimate the compatibility score between a property $\mathbf{c}$ and data point $\mathbf{x}$ space dot product:
\begin{equation}
    \mathrm{score}(c,x) = \Phi(c)^{T}\Phi(x),
\end{equation}
where a higher score indicates the corresponding property and data points are more likely to be linked.

To learn the parameter weights of the encoder, we fine-tune an off-the-shelf dense retriever model\footnote{sentence-transformers/all-MiniLM-L6-v2} using a batch-wise-negative-sampling-based loss function~\cite{henderson2017efficientnaturallanguageresponse} to predict whether a property $c$ is generated from a data point $x$ in the property proposal stage.
Specifically, for any given data point in the training dataset, our framework estimates the probability of its corresponding property as 
\begin{equation}
    \mathrm{p}(c|x) = \frac{\exp(\mathrm{score}(c|x))}{\sum_{j=1}^{K} \exp(\mathrm{score}(c|x))},
\end{equation}
where we consider $K-1$ randomly sampled negative samples given a positive property-data point pair.
This negative sampling scheme assumes randomly sampled pairs are most frequently incompatible pairs, which is a common assumption in matrix factorization applications such as recommender systems~\cite{he2017neuralCF}.

After learning the encoder, we can efficiently perform goal-oriented data transformation $f$ as a linear transformation of the input data:
\begin{equation}
     \mathrm{W}^{T}\Phi(\mathrm{x}), \text{where } \mathrm{W}_{i} = \Phi(\mathrm{c_i}),
\end{equation}
where the weight matrix $\mathrm{W}$ is a pre-computed matrix where each row is the representation for a property.
The output of this operation is a matrix, in which each value represents an estimated compatibility score between a property and a data point.
Next, we group properties with high covariance in the matrix into clusters representing latent factors.

\subsection{Latent Factor Discovery}

Properties generated by LLMs are naturally noisy, and can contain duplicate or highly correlated phrases.
Without further processing, these raw properties would result in an overly complex system that is hard to interpret, similar to how each individual word in a topic model is uninformative about latent patterns in a dataset~\cite{Blei2009LatentDA, Grootendorst2022BERTopicNT}.
In other words, these properties need to be grouped into higher-level patterns before they can be of use to end users.
To this end, we propose to cluster the properties by their correlations in the estimated compatibility matrix.
Thus, we propose to treat the proposed properties $C$ as observed variables, and seek to infer higher-level latent patterns $Z$ from their estimated compatibility score matrix from the goal-oriented data transformation step.

There are various model choices for learning latent variables from a set of properties from a tabular dataset.
However, another challenge in this case is we want to not only learn a model of fit, but also cluster potentially large quantities of correlated properties.
To this end, we adopt Linear Corex, a state-of-the-art model in latent structure learning~\cite{Steeg19linearcorex} that scales well with input dimensionality, and cast this clustering problem into learning a modular latent factor model over the (gaussianized compatibility scores of) properties that aims to satisfy the following condition: $\mathit{TC}(C|Z) + TC(Z) = 0$ and $\forall i, \mathit{TC}(Z|C_i) = 0$,
where $\mathrm{TC}$ stands for the total correlation for a random variable:
\begin{equation}
    \mathit{TC}(Y) = \sum_{i=1}^{N}H(Y_i) - H(Y).
\end{equation}
In this case, $H$ denotes differential entropy.
This formulation encourages each property to be assigned to only one latent dimension in $Z$ via the modular $\mathit{TC}(Z|C_i)=0$ constraint, and are thus suited for our goal of clustering the properties.

To this end, we fit a linear latent factor model with the loss function proposed by~\citeauthor{Steeg19linearcorex} to encourage the solution to better align with conditions discussed above (see~\Cref{sec:detail_for_steeg} for details).
We can then group properties assigned to (i.e. has high mutual information with) the same latent concept into a discovered latent factor.
Importantly, since each of our property is associated with a natural language description, the grouped set of properties provides interpretability of the discovered latent factor.

\section{Problem Setup}
\label{sec:problem_setup}

Quantitatively evaluating task-oriented latent factor discovery is challenging in that there is not always an intuitive method to measure whether the discovered latent factors are truly informative and related to the users' discovery goal.
To this end, we select three use cases where the task usefulness of the discovered latent factors can indeed be quantitatively evaluated with task-specific evaluations.
Specifically, we adopt popular choice of evaluation in representation learning~\cite{ Nozawa2022EvaluationMF}, and use performance derived from latent representations on downstream tasks as a proxy for evaluating the latent space of latent factor discovery models.
The trends from automated evaluation are then corroborated via user studies.

In particular, we experiment with (1) discovering factors related to users' interest in movies from conversational recommendation dialogues, (2) discovering factors related to users' actions from embodied navigation action logs, and (3) discovering factors related to document topics from a set of documents. 
Each of these scenarios embeds a goal whose success can be quantitatively evaluated via downstream predictive tasks, namely conversational recommendation (\Cref{sec:movie_recommendation}), user action prediction (\Cref{sec:embodied_navigation}), and document labeling (\Cref{sec:document_categorization}), which helps to verify the effectiveness of latent factor discovery systems.
We provide detailed discussion on task-specific evaluations in experiment sections.

\section{Experiments}

\paragraph{Baselines}
We compare our method with (1) classic gradient-based baselines (LDA~\cite{Blei2009LatentDA} and BERTopic~\cite{Grootendorst2022BERTopicNT}), and (2) TopicGPT~\cite{pham-etal-2024-topicgpt}, a recent state-of-the-art LLM-reasoning-based framework for topic-modeling.
We adapt TopicGPT by additionally including users' goal for discovery into its prompts.
By default, TopicGPT use early-stopping that stops the topic proposal process when a pre-defined number of duplicate concepts (100 in practice) are generated.
To this end, we also experiment with another variant with such constraint lifted, denoted by TopicGPT-full.

While not all these frameworks are designed to take user goal into account, they represent the most applicable current method for uncovering goal-oriented latent factors from unstructured data.
Other than the default setting for~\ourmethod, we additionally evaluate a binarized version of our model,~\ourmethod-BIN, where the estimated compatibility score in $C$ are binary.
This model variant represents the extreme case where the user wants to have full interpretabilty on whether certain property is related to a data point as a binary value.
For all LLM-based methods, we evaluate with two representative LLMs: GPT-3.5 and GPT-4o.
To evaluate whether~\ourmethod~is indeed less reliant on a strong base LLM, we also report performance using Mistral-7b~\cite{jiang2023mistral7b}, an open-source language model which prior work find cannot produce coherent topics due to noisy generation results~\cite{pham-etal-2024-topicgpt}.
We discuss hyper-parameters (\Cref{sec:implementation_details}), efficiency (\Cref{sec:discussion_on_efficiency}), case studies (\Cref{sec:case_studies}) and prompt stability (\Cref{sec:discussion_on_prompts}) in the appendix.

\section{Movie Recommendation}
\label{sec:movie_recommendation}

\begin{table}[h!]
\small
    \centering
    \resizebox{\columnwidth}{!}{%
    \begin{tabular}{lrrr}
        \toprule
        Model & H@1 & H@5 & H@20 \\
        \midrule
        Majority & 4.32 & 9.13 & 21.15 \\
        LDA & 0.96 & 0.96 & 1.92 \\
        BERTopic & 1.92 & 1.92 & 2.88 \\
        TopicGPT-3.5 & 1.90 & 2.40 & 2.80 \\
        TopicGPT-3.5-full & 2.40 & 3.36 & 3.80 \\
        TopicGPT-4o & 0.48 & 0.48 & 1.44 \\
        TopicGPT-4o-full & 1.92 & 1.92 & 2.88 \\
        \midrule
        \ourmethod-Mistral & \textbf{4.80} & 11.53 & \textbf{24.03} \\
        \ourmethod-3.5-BIN & 1.90 & 10.50 & \underline{23.00} \\
        \ourmethod-3.5 & \underline{4.30} & \textbf{13.90} & 23.50 \\
        \ourmethod-4o & 3.84 & \underline{12.90} & 20.60 \\
        \bottomrule
    \end{tabular}
    }
    \caption{Performance (Hit at k) on movie recommendation on the Inspired dataset.}
    \label{tab:movie_recommendation_metrics}
\vspace{0mm}
\end{table}

On conversational recommendation (CR) task, we use the Inspired~\cite{hayati-etal-2020-inspired} dataset, a widely used CR dataset with semantically rich multi-turn dialogues.
We follow the same dataset split procedures as in prior works~\cite{xie2024neighbor, he23large}, randomly partitioning the dataset into 731 training and 211 test samples. 
We provide additional dataset details in~\Cref{sec:implementation_details}.
To compare the quality of different methods on the dataset, we adopt NBCRS~\cite{xie2024neighbor}, a retrieval-based conversational recommender system that makes recommendations by outputting popular movies in a semantic neighborhood with any document representation methods.

Concretely, given a test dialogue history (i.e., previous interactions between the user and the assistant) and a method that is being tested, we encode the history into embedding using the latent space discovered by the corresponding method, and retrieve its $k$ nearest neighbor in the embedding space.
Following recent works on conversational recommendation~\cite{xie2024neighbor, he23large}, we evaluate the performance of systems by Hit@$k$ w.r.t. the ground-truth movie mentioned in the response to the dialogue history, where $k \in {1, 5, 20}$.
The performance of the models is as shown in~\Cref{tab:movie_recommendation_metrics}.
As shown in the table, our method is the only latent factor model that can meaningfully organize data points in the latent space and has good performance.
Notably, our method still performs well when the model is binarized.
This shows that our methods discover task-relevant and informative latent properties from data.

\section{Embodied Navigation on Alfworld}
\label{sec:embodied_navigation}

\begin{table}[h!]
\small
    \centering
    \resizebox{\columnwidth}{!}{%
    \begin{tabular}{lrr}
        \toprule
        Model & Seen Task & Unseen Task \\
        \midrule
        Majority & 5.60 & 8.67 \\
        LDA & 33.1 & 17.34 \\
        BERTopic & 20.30 & 22.47 \\
        TopicGPT-3.5 & 1.69 & 0.65 \\
        TopicGPT-3.5-full & 4.30 & 5.08 \\
        TopicGPT-4o & 1.04 & 5.12 \\
        TopicGPT-4o-full & 1.43 & 2.36 \\
        \midrule
        \ourmethod-Mistral & \underline{48.63} & 32.45 \\
        \ourmethod-3.5-BIN & 45.37 & 33.24 \\
        \ourmethod-3.5 & 48.10 & \underline{33.77} \\
        \ourmethod-4o & \textbf{49.28} & \textbf{34.42} \\
        \bottomrule
    \end{tabular}
    }
    \caption{Performance on next-action prediction for embodied navigation on Alfworld.}
    \label{tab:embodied_navigation_metrics}
    \vspace{0mm}
\end{table}

For the second scenario, we evaluate the performance of our method on next action prediction on Alfworld~\cite{shridhar2021alfworld} navigation logs.
Specifically, given a user interaction log with the Alfworld environment in the test set, we retrieve the most similar training interaction log in the latent space from the training set using cosine similarity, and then check whether the next action associated with the training interaction log is the same as the test-time ground truth.

We chose next action prediction instead of directly aiming for a higher score on Alfworld since this is a direct evaluation to test latent representation quality than a whole systems' end performance.
This is similar to how model-based probing method requires simple models rather than complex systems to check latent representation quality in prior works~\cite{Nozawa2022EvaluationMF}.
We provide further discussions on this in~\Cref{sec:task_diffuculties}.
To create the test dataset, we take trajectories from both the Seen tasks and Unseen tasks categories from the test set of Alfworld, and break the trajectories into context-and-next-action pairs.
I.e., given a trajectory sequence $<s_1, a_1, s_2, a_2, ..., s_n, a_n>$, any $<s_1, a_1, ..., s_i>$ is a valid state, where the next action is $a_i$. 
Since the interaction logs on Alfworld are purely text-based, the whole trajectory then becomes an unstructured data point in the context of~\ourmethod.

The performance of~\ourmethod~against baselines is as shown in~\Cref{tab:embodied_navigation_metrics}.
Our method outperforms the baseline method, and notably, has minimal degradation when the compatibility score between a data point and a property is binarized.
We hypothesize that this is due to properties on Alfworld are often highly un-ambiguous (e.g., \texttt{``the user is tasked with cleaning an item"}), and thus, in this case, binarized compatibility scores are already sufficiently expressive.
On the other hand, we observe that due to the reliance on an LLMs' ability to handle both topic generation and assignment, TopicGPT has degraded performance on this task.
We hypothesize that this is due to TopicGPT being designed for topic modeling, where the documents to be categorized are often in the pre-training distribution of an LLM, versus~\ourmethod~does not rely on LLMs in proposing high-level topics, thus are less reliant on LLMs' knowledge, and is thus more robust to the noisy, out-of-distribution interaction logs on Alfworld.

\section{The American Bills Dataset}
\label{sec:document_categorization}

\begin{table}[h!]
    \centering
    \small
    \resizebox{\columnwidth}{!}{%
    \begin{tabular}{lrr}
        \toprule
        Model & High-level & Fine grained \\
        \midrule
        Majority & 11.57 & 5.90 \\
        LDA & 40.72 & 19.52 \\
        BERTopic & \underline{52.94} & 27.55 \\
        TopicGPT-3.5 & 51.18 & 19.11 \\
        TopicGPT-3.5-full &\textbf{55.76} & 22.59 \\
        TopicGPT-4o & 51.14 & 19.13 \\
        TopicGPT-4o-full & 49.40 & 14.49 \\
        \midrule
        \ourmethod-Mistral & 49.29 & 28.46 \\
        \ourmethod-3.5-BIN & 47.11 & 25.25 \\
        \ourmethod-3.5 & 51.50 & \textbf{31.09} \\
        \ourmethod-4o & 52.40 & \underline{29.30} \\
        \bottomrule
    \end{tabular}
    }
    \caption{Performance on document categorization on the Bills dataset.}
    \label{tab:document_categorization_metrics}
    \vspace{0mm}
\end{table}

Finally, we experiment with the American Bills dataset~\cite{hoyle-etal-2022-neural}. 
We pick this dataset since this is a widely-used 
dataset for topic modeling, and is the only dataset for which the prompt for our LLM-based baseline~\cite{pham-etal-2024-topicgpt} is publically available.
To this end, the purpose of this dataset is to show that under a more classical setting where the users' goal is obvious from the dataset, our method still performs well compared to the baselines.
The original dataset subset in~\citeauthor{pham-etal-2024-topicgpt} contains 16,242 American bill summaries.
To ensure there is a sufficient number of documents in both the training and evaluation set, we moved half of the original evaluation set into the training set at random, resulting in a dataset of 8981 training data points and 7261 evaluation data points.

To evaluate the quality of our latent-factor-discovery method against baselines, we apply decision-tree-based probing to see if we can derive both high-level and fine-grained labels in the dataset by training a classifier on the representation produced by~\ourmethod~and baseline methods on the evaluation set.
We report the average class-balanced accuracy score from five-fold cross-validation (training the decision tree only) for deriving both high-level and fine-grained labels using latent representations produced by each baseline.

The performance of the methods is as shown in table~\ref {tab:document_categorization_metrics}. 
All methods except LDA have comparable performance on recovering high-level topics.
However, our method outperforms the baseline methods for recovering fine-grained topics, showing that~\ourmethod~can discover informative latent factors even on a traditional dataset that suits topic models well.
Meanwhile, we observe that TopicGPT and BERTopic also demonstrate competitive performance in this case, showing that prior methods are still a viable solution for uncovering hidden topics from unstructured documents that clearly exhibit meaningful topics.

\section{Human Evaluation and Ablations}
\label{sec:human_eval_and_ablations}

\begin{table*}[t!]
\small
\centering
\setlength\tabcolsep{4pt}
\begin{tabular}{l | ccc c ccc c ccc}
\toprule
\multirow{2}{*}{\textbf{Number of Wins}}
& \multicolumn{3}{c}{\textbf{Task Relevance}} &
& \multicolumn{3}{c}{\textbf{Informativeness}} &
& \multicolumn{3}{c}{\textbf{Overall}} \\
\cmidrule(lr){2-4} \cmidrule(lr){6-8} \cmidrule(lr){10-12}
& \textbf{Inspired} & \textbf{Alfworld} & \textbf{Bills} &
& \textbf{Inspired} & \textbf{Alfworld} & \textbf{Bills} &
& \textbf{Inspired} & \textbf{Alfworld} & \textbf{Bills} \\
\midrule
LDA & \underline{15} & \underline{16} & \underline{26} &
    & 6 & \underline{6} & 9 &
    & \underline{13} & \underline{12} & 18 \\
BERTopic & 13 & 11 & 20 &
    & 9 & \underline{6} & \underline{13} &
    & 9 & \underline{12} & \underline{22} \\
TopicGPT & 11 & 3 & 5 &
    & \underline{18} & 3 & 1 &
    & 10 & 3 & 2 \\
\ourmethod & \textbf{26}\textsuperscript{*} & \textbf{32}\textsuperscript{*} & \textbf{28} &
    & \textbf{23} & \textbf{45}\textsuperscript{*} & \textbf{33}\textsuperscript{*} &
    & \textbf{26}\textsuperscript{*} & \textbf{37} & \textbf{26} \\
\bottomrule
\end{tabular}%
\caption{Human evaluation results: number of wins each method had in 50 head-to-head comparisons where users select multiple best methods. An asterisk (*) indicates methods that are statistically significantly better than the second-best with $p < 0.05$.}
\vspace{-1em}
\label{tab:human_relevance_informativeness}
\end{table*}

\paragraph{Task-relevance and Informativeness}

While we quantitatively demonstrate the task-effectiveness of our framework with automated evaluation, a core requirement of a successful latent-factor discovery system is that it should uncover informative, task-relevant features and present the findings via interpretable signals to the user.
To this end, we run a human evaluation on Amazon Mechanical Turk with the question of whether~\ourmethod~can uncover good-quality latent properties.
To assess the overall performance of different frameworks, we also ask human evaluators examine a discovered factor from all methods, and pick out one or more frameworks they prefer to use most when tasked to understand patterns in a dataset (see~\Cref{sec:human_evaluation_details} for details).
The results, as shown in~\Cref{tab:human_relevance_informativeness}, confirm the effectiveness of our method. It outperforms the baseline method in terms of task-relevance and informativeness, and is consistently rated as the overall best model in human evaluations, reinforcing its effectiveness and user preference. This corroborates the automated evaluation results.

\begin{table}[t!]
\small
\setlength\tabcolsep{11pt}
\center
\begin{tabular}{lrrr}
\toprule
{} &  \bf Insp. & \bf Alf. &  \bf Bills  \\
\midrule
BERTopic  & 5.0  & 3.86 &  3.62  \\
\ourmethod  &  \textbf{1.66} & \textbf{1.18} &  \textbf{1.62}  \\
\bottomrule
\end{tabular}
\caption{Average number of outlier property @ 10, lower number indicates better performance. Differences are significant with $p<0.05$.}
\label{tab:observation_grouping_qual}
\vspace{-3mm}
\end{table}

\paragraph{Quality of property grouping}

We are additionally interested in evaluating whether the grouping in~\ourmethod~can correctly assign meaningfully correlated properties into the same latent factor.
To this end, we randomly sample 10 properties from our framework and ask the human evaluator first to identify a topic from the keywords or phrases, then report the number of outliers in the samples.
To establish a baseline for this evaluation setting, we pick BERTopic, the best-performing baseline that groups keywords into properties, and select the top 10 keywords for each topic it discovers. 
As shown in~\Cref{tab:observation_grouping_qual}, \ourmethod~can produce coherent clusters of fine-grained properties, in that human evaluators find fewer outliers in the latent factor interpretations identified by our framework.

\paragraph{Why not prompt LLMs to link properties and data points?} 
\label{sec:llms_are_better_generator}
In this section, we show LLMs are better at generating properties than assigning properties to data points.
Specifically, recent studies show generative (language) models do not understand their own-generated contents well~\cite{Qiu2023PhenomenalYP}.
To this end, we demonstrate that this phenomena is also true in the context of using LLM to link properties to data.

\begin{figure}[htb]
\centering
\includegraphics[scale=0.6]{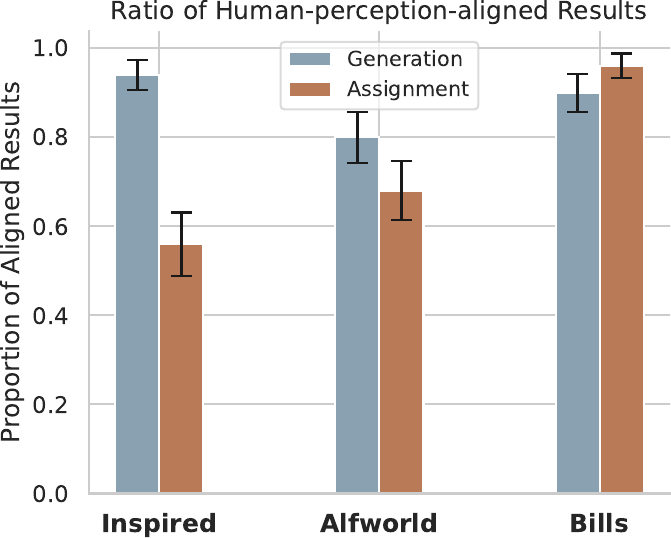}
\caption{LLMs are better property proposers than generators for GPT-3.5. See~\Cref{fig:llm_are_better_generator_than_assigner_4o} for results on GPT-4o, where the trend is consistent.
}
\label{fig:llm_are_better_generator_than_assigner}
\vskip -5mm
\end{figure}

We conduct a human evaluation to compare LLMs' ability to generate property from a data point versus the their ability to predict the entailment between a property and a data point by answering ``yes" or ``no".
In particular, after generating the properties on each scenario, we prompt LLM to predict whether a property applies to a data point and collect a set of data-property pairs linked via LLM assignment. 
We then ask human evaluators to judge the validity of data-property pairs linked from both property proposal and assignment on a 5-score Likert scale, where a score of 4 and 5 indicates ``likely correct" and ``absolutely correct", with the remaining scores denoting ``neutral" or worse.
We report the number of ``likely correct" or beyond out of 50 trials in~\Cref{tab:observation_grouping_qual}.
As shown, LLMs' ability to assign properties to data points lags behind its ability to generate properties.
We hypothesize that this is because generating descriptive properties based on an input document is more common in LLMs' pre-train corpus than property assignments, where a decision of entailment naturally follows a document and a set of properties.

\section{Conclusion and Future Works}

We develop~\ourmethod, a latent factor discovery framework that uncovers task-relevant, informative, and interpretable latent concepts from unstructured data based on users' instruction in natural language.
\ourmethod~combines the instruction-following ability of LLMs and the scalability of gradient-based latent factor models, demonstrating the promise of improving statistical algorithms with LLM reasoning ability.

\section{Limitations}

Similar to a body of recent works, our frameworks require an LLM with reasoning ability, and we opt to evaluate our method on two widely used close-source models and an open-source model following prior works on LLM reasoning~\cite{yu-etal-2023-prompt, wang2024promptagent, nottingham2023do, Qiu2023PhenomenalYP}.
While our framework relieves in reliance on a strong LLM than baseline~\cite{pham-etal-2024-topicgpt}, it would be an interesting direction to explore future frameworks that can work well with a smaller generative model, such as a fine-tuned T5~\cite{Raffel2019ExploringTL} model.
Finally, we note that dot product is not the only viable option for estimating the compatibility scores between documents and properties, and leave exploration to other alternatives such as cosine similarities to future works.

\section{Potential Risks and Ethical Concerns}

We note that LLM are known to suffer from hallucinations in its generated content~\cite{zhao2024surveylargelanguagemodels}.
To this end, we advise practitioners to carefully verify the generated content from our framework before deploying it in critical decision-making scenarios.

\section*{Acknowledgement}

We thank the anonymous reviewers, the Aristo and the Semantic Scholar team, Dongfu Jiang, Bill Yuchen Lin, and other members at the Allen Institute for AI for their insightful comments.

\bibliography{references}

\begin{thebibliography}{40}
\expandafter\ifx\csname natexlab\endcsname\relax\def\natexlab#1{#1}\fi

\bibitem[{Blei et~al.(2009)Blei, Ng, and Jordan}]{Blei2009LatentDA}
David~M. Blei, A.~Ng, and Michael~I. Jordan. 2009.
\newblock \href {https://api.semanticscholar.org/CorpusID:3177797} {Latent dirichlet allocation}.

\bibitem[{Bragg et~al.(2013)Bragg, Mausam, and Weld}]{Bragg2013CrowdsourcingMC}
Jonathan Bragg, Mausam, and Daniel~S. Weld. 2013.
\newblock \href {https://api.semanticscholar.org/CorpusID:11974604} {Crowdsourcing multi-label classification for taxonomy creation}.
\newblock In \emph{AAAI Conference on Human Computation \& Crowdsourcing}.

\bibitem[{Fayyad et~al.(1996)Fayyad, Piatetsky-Shapiro, and Smyth}]{Fayyad1996TheKP}
Usama~M. Fayyad, Gregory Piatetsky-Shapiro, and Padhraic Smyth. 1996.
\newblock \href {https://api.semanticscholar.org/CorpusID:56635038} {The kdd process for extracting useful knowledge from volumes of data}.
\newblock \emph{Commun. ACM}, 39:27--34.

\bibitem[{Grootendorst(2022)}]{Grootendorst2022BERTopicNT}
Maarten~R. Grootendorst. 2022.
\newblock \href {https://api.semanticscholar.org/CorpusID:247411231} {Bertopic: Neural topic modeling with a class-based tf-idf procedure}.
\newblock \emph{ArXiv}, abs/2203.05794.

\bibitem[{Hayati et~al.(2020)Hayati, Kang, Zhu, Shi, and Yu}]{hayati-etal-2020-inspired}
Shirley~Anugrah Hayati, Dongyeop Kang, Qingxiaoyang Zhu, Weiyan Shi, and Zhou Yu. 2020.
\newblock \href {https://doi.org/10.18653/v1/2020.emnlp-main.654} {{INSPIRED}: Toward sociable recommendation dialog systems}.
\newblock In \emph{Proceedings of the 2020 Conference on Empirical Methods in Natural Language Processing (EMNLP)}, pages 8142--8152, Online. Association for Computational Linguistics.

\bibitem[{He et~al.(2017)He, Liao, Zhang, Nie, Hu, and Chua}]{he2017neuralCF}
Xiangnan He, Lizi Liao, Hanwang Zhang, Liqiang Nie, Xia Hu, and Tat-Seng Chua. 2017.
\newblock Neural collaborative filtering.
\newblock In \emph{WWW}, WWW '17, Republic and Canton of Geneva, CHE. International World Wide Web Conferences Steering Committee.

\bibitem[{He et~al.(2023)He, Xie, Jha, Steck, Liang, Feng, Majumder, Kallus, and McAuley}]{he23large}
Zhankui He, Zhouhang Xie, Rahul Jha, Harald Steck, Dawen Liang, Yesu Feng, Bodhisattwa Majumder, Nathan Kallus, and Julian McAuley. 2023.
\newblock Large language models as zero-shot conversational recommenders.
\newblock In \emph{CIKM}.

\bibitem[{Henderson et~al.(2017)Henderson, Al-Rfou, Strope, hsuan Sung, Lukacs, Guo, Kumar, Miklos, and Kurzweil}]{henderson2017efficientnaturallanguageresponse}
Matthew Henderson, Rami Al-Rfou, Brian Strope, Yun hsuan Sung, Laszlo Lukacs, Ruiqi Guo, Sanjiv Kumar, Balint Miklos, and Ray Kurzweil. 2017.
\newblock \href {http://arxiv.org/abs/1705.00652} {Efficient natural language response suggestion for smart reply}.

\bibitem[{Hoyle et~al.(2022)Hoyle, Sarkar, Goel, and Resnik}]{hoyle-etal-2022-neural}
Alexander~Miserlis Hoyle, Rupak Sarkar, Pranav Goel, and Philip Resnik. 2022.
\newblock \href {https://doi.org/10.18653/v1/2022.findings-emnlp.390} {Are neural topic models broken?}
\newblock In \emph{Findings of the Association for Computational Linguistics: EMNLP 2022}, pages 5321--5344, Abu Dhabi, United Arab Emirates. Association for Computational Linguistics.

\bibitem[{Jiang et~al.(2023)Jiang, Sablayrolles, Mensch, Bamford, Chaplot, de~las Casas, Bressand, Lengyel, Lample, Saulnier, Lavaud, Lachaux, Stock, Scao, Lavril, Wang, Lacroix, and Sayed}]{jiang2023mistral7b}
Albert~Q. Jiang, Alexandre Sablayrolles, Arthur Mensch, Chris Bamford, Devendra~Singh Chaplot, Diego de~las Casas, Florian Bressand, Gianna Lengyel, Guillaume Lample, Lucile Saulnier, Lélio~Renard Lavaud, Marie-Anne Lachaux, Pierre Stock, Teven~Le Scao, Thibaut Lavril, Thomas Wang, Timothée Lacroix, and William~El Sayed. 2023.
\newblock \href {http://arxiv.org/abs/2310.06825} {Mistral 7b}.

\bibitem[{Jolliffe and Cadima(2016)}]{Jolliffe2016PrincipalCA}
Ian~T. Jolliffe and Jorge Cadima. 2016.
\newblock \href {https://api.semanticscholar.org/CorpusID:20101754} {Principal component analysis: a review and recent developments}.
\newblock \emph{Philosophical Transactions of the Royal Society A: Mathematical, Physical and Engineering Sciences}, 374.

\bibitem[{Kang et~al.(2018)Kang, Ammar, Dalvi, van Zuylen, Kohlmeier, Hovy, and Schwartz}]{kang-etal-2018-dataset}
Dongyeop Kang, Waleed Ammar, Bhavana Dalvi, Madeleine van Zuylen, Sebastian Kohlmeier, Eduard Hovy, and Roy Schwartz. 2018.
\newblock \href {https://doi.org/10.18653/v1/N18-1149} {A dataset of peer reviews ({P}eer{R}ead): Collection, insights and {NLP} applications}.
\newblock In \emph{Proceedings of the 2018 Conference of the North {A}merican Chapter of the Association for Computational Linguistics: Human Language Technologies, Volume 1 (Long Papers)}, pages 1647--1661, New Orleans, Louisiana. Association for Computational Linguistics.

\bibitem[{Kingma and Welling(2013)}]{Kingma2013AutoEncodingVB}
Diederik~P. Kingma and Max Welling. 2013.
\newblock \href {https://api.semanticscholar.org/CorpusID:216078090} {Auto-encoding variational bayes}.
\newblock \emph{CoRR}, abs/1312.6114.

\bibitem[{Koh et~al.(2020)Koh, Nguyen, Tang, Mussmann, Pierson, Kim, and Liang}]{koh2020conceptbottleneckmodels}
Pang~Wei Koh, Thao Nguyen, Yew~Siang Tang, Stephen Mussmann, Emma Pierson, Been Kim, and Percy Liang. 2020.
\newblock \href {http://arxiv.org/abs/2007.04612} {Concept bottleneck models}.

\bibitem[{Liu et~al.(2016)Liu, Tang, Dong, Yao, and Zhou}]{liu2016overviewtopicbio}
Lin Liu, Lin Tang, Wen Dong, Shaowen Yao, and Wei Zhou. 2016.
\newblock An overview of topic modeling and its current applications in bioinformatics.
\newblock \emph{SpringerPlus}, 5:1--22.

\bibitem[{McAuley and Leskovec(2013)}]{mcauley2013hidden}
Julian McAuley and Jure Leskovec. 2013.
\newblock \href {https://doi.org/10.1145/2507157.2507163} {Hidden factors and hidden topics: understanding rating dimensions with review text}.
\newblock In \emph{Proceedings of the 7th ACM Conference on Recommender Systems}, RecSys '13, page 165–172, New York, NY, USA. Association for Computing Machinery.

\bibitem[{McAuley et~al.(2015)McAuley, Pandey, and Leskovec}]{mcauley2015inferring}
Julian McAuley, Rahul Pandey, and Jure Leskovec. 2015.
\newblock \href {https://doi.org/10.1145/2783258.2783381} {Inferring networks of substitutable and complementary products}.
\newblock In \emph{Proceedings of the 21th ACM SIGKDD International Conference on Knowledge Discovery and Data Mining}, KDD '15, page 785–794, New York, NY, USA. Association for Computing Machinery.

\bibitem[{Nottingham et~al.(2023)Nottingham, Ammanabrolu, Suhr, Choi, Hajishirzi, Singh, and Fox}]{nottingham2023do}
Kolby Nottingham, Prithviraj Ammanabrolu, Alane Suhr, Yejin Choi, Hannaneh Hajishirzi, Sameer Singh, and Roy Fox. 2023.
\newblock \href {https://openreview.net/forum?id=Z_qiOvqvnBl} {Do embodied agents dream of pixelated sheep?: Embodied decision making using language guided world modelling}.
\newblock In \emph{Workshop on Reincarnating Reinforcement Learning at ICLR 2023}.

\bibitem[{Nozawa and Sato(2022)}]{Nozawa2022EvaluationMF}
Kento Nozawa and Issei Sato. 2022.
\newblock \href {https://api.semanticscholar.org/CorpusID:250634248} {Evaluation methods for representation learning: A survey}.
\newblock In \emph{International Joint Conference on Artificial Intelligence}.

\bibitem[{Oikarinen et~al.(2023{\natexlab{a}})Oikarinen, Das, Nguyen, and Weng}]{oikarinen2023labelfree}
Tuomas Oikarinen, Subhro Das, Lam~M. Nguyen, and Tsui-Wei Weng. 2023{\natexlab{a}}.
\newblock \href {https://openreview.net/forum?id=FlCg47MNvBA} {Label-free concept bottleneck models}.
\newblock In \emph{The Eleventh International Conference on Learning Representations}.

\bibitem[{Oikarinen et~al.(2023{\natexlab{b}})Oikarinen, Das, Nguyen, and Weng}]{oikarinen2023labelfreeconceptbottleneckmodels}
Tuomas Oikarinen, Subhro Das, Lam~M. Nguyen, and Tsui-Wei Weng. 2023{\natexlab{b}}.
\newblock \href {http://arxiv.org/abs/2304.06129} {Label-free concept bottleneck models}.

\bibitem[{Pham et~al.(2024)Pham, Hoyle, Sun, Resnik, and Iyyer}]{pham-etal-2024-topicgpt}
Chau Pham, Alexander Hoyle, Simeng Sun, Philip Resnik, and Mohit Iyyer. 2024.
\newblock \href {https://doi.org/10.18653/v1/2024.naacl-long.164} {{T}opic{GPT}: A prompt-based topic modeling framework}.
\newblock In \emph{Proceedings of the 2024 Conference of the North American Chapter of the Association for Computational Linguistics: Human Language Technologies (Volume 1: Long Papers)}, pages 2956--2984, Mexico City, Mexico. Association for Computational Linguistics.

\bibitem[{Qiu et~al.(2024)Qiu, Jiang, Lu, Sclar, Pyatkin, Bhagavatula, Wang, Kim, Choi, Dziri, and Ren}]{Qiu2023PhenomenalYP}
Linlu Qiu, Liwei Jiang, Ximing Lu, Melanie Sclar, Valentina Pyatkin, Chandra Bhagavatula, Bailin Wang, Yoon Kim, Yejin Choi, Nouha Dziri, and Xiang Ren. 2024.
\newblock \href {https://api.semanticscholar.org/CorpusID:263909078} {Phenomenal yet puzzling: Testing inductive reasoning capabilities of language models with hypothesis refinement}.
\newblock \emph{ICLR}.

\bibitem[{Raffel et~al.(2019)Raffel, Shazeer, Roberts, Lee, Narang, Matena, Zhou, Li, and Liu}]{Raffel2019ExploringTL}
Colin Raffel, Noam~M. Shazeer, Adam Roberts, Katherine Lee, Sharan Narang, Michael Matena, Yanqi Zhou, Wei Li, and Peter~J. Liu. 2019.
\newblock \href {https://api.semanticscholar.org/CorpusID:204838007} {Exploring the limits of transfer learning with a unified text-to-text transformer}.
\newblock \emph{J. Mach. Learn. Res.}, 21:140:1--140:67.

\bibitem[{Ramage et~al.(2009)Ramage, Rosen, Chuang, Manning, and McFarland}]{ramage2009topic}
Daniel Ramage, Evan Rosen, Jason Chuang, Christopher~D Manning, and Daniel~A McFarland. 2009.
\newblock Topic modeling for the social sciences.
\newblock In \emph{NIPS 2009 workshop on applications for topic models: text and beyond}, volume~5, pages 1--4.

\bibitem[{Rumelhart et~al.(1986)Rumelhart, Hinton, and Williams}]{Rumelhart1986LearningIR}
David~E. Rumelhart, Geoffrey~E. Hinton, and Ronald~J. Williams. 1986.
\newblock \href {https://api.semanticscholar.org/CorpusID:62245742} {Learning internal representations by error propagation}.

\bibitem[{Shridhar et~al.(2021)Shridhar, Yuan, Cote, Bisk, Trischler, and Hausknecht}]{shridhar2021alfworld}
Mohit Shridhar, Xingdi Yuan, Marc-Alexandre Cote, Yonatan Bisk, Adam Trischler, and Matthew Hausknecht. 2021.
\newblock \href {https://openreview.net/forum?id=0IOX0YcCdTn} {{\{}ALFW{\}}orld: Aligning text and embodied environments for interactive learning}.
\newblock In \emph{International Conference on Learning Representations}.

\bibitem[{Steeg et~al.(2017)Steeg, Harutyunyan, Moyer, and Galstyan}]{Steeg2017FastSL}
Greg~Ver Steeg, Hrayr Harutyunyan, Daniel Moyer, and A.~G. Galstyan. 2017.
\newblock \href {https://api.semanticscholar.org/CorpusID:202539161} {Fast structure learning with modular regularization}.
\newblock In \emph{Neural Information Processing Systems}.

\bibitem[{Steeg et~al.(2019)Steeg, Harutyunyan, Moyer, and Galstyan}]{Steeg19linearcorex}
Greg~Ver Steeg, Hrayr Harutyunyan, Daniel Moyer, and Aram Galstyan. 2019.
\newblock \href {https://proceedings.neurips.cc/paper/2019/hash/e2e14235335d2c0aa5f6855e339233d9-Abstract.html} {Fast structure learning with modular regularization}.
\newblock In \emph{Advances in Neural Information Processing Systems 32: Annual Conference on Neural Information Processing Systems 2019, NeurIPS 2019, December 8-14, 2019, Vancouver, BC, Canada}, pages 15567--15577.

\bibitem[{Wan et~al.(2024)Wan, Safavi, Jauhar, Kim, Counts, Neville, Suri, Shah, White, Yang, Andersen, Buscher, Joshi, and Rangan}]{wan2024tntlm}
Mengting Wan, Tara Safavi, Sujay~Kumar Jauhar, Yujin Kim, Scott Counts, Jennifer Neville, Siddharth Suri, Chirag Shah, Ryen~W. White, Longqi Yang, Reid Andersen, Georg Buscher, Dhruv Joshi, and Nagu Rangan. 2024.
\newblock \href {https://doi.org/10.1145/3637528.3671647} {Tnt-llm: Text mining at scale with large language models}.
\newblock In \emph{Proceedings of the 30th ACM SIGKDD Conference on Knowledge Discovery and Data Mining}, KDD '24, page 5836–5847, New York, NY, USA. Association for Computing Machinery.

\bibitem[{Wang et~al.(2024)Wang, Li, Wang, Bai, Luo, Zhang, Jojic, Xing, and Hu}]{wang2024promptagent}
Xinyuan Wang, Chenxi Li, Zhen Wang, Fan Bai, Haotian Luo, Jiayou Zhang, Nebojsa Jojic, Eric Xing, and Zhiting Hu. 2024.
\newblock \href {https://openreview.net/forum?id=22pyNMuIoa} {Promptagent: Strategic planning with language models enables expert-level prompt optimization}.
\newblock In \emph{The Twelfth International Conference on Learning Representations}.

\bibitem[{Wang et~al.(2023)Wang, Shang, and Zhong}]{wang-etal-2023-goal}
Zihan Wang, Jingbo Shang, and Ruiqi Zhong. 2023.
\newblock \href {https://doi.org/10.18653/v1/2023.emnlp-main.657} {Goal-driven explainable clustering via language descriptions}.
\newblock In \emph{Proceedings of the 2023 Conference on Empirical Methods in Natural Language Processing}, pages 10626--10649, Singapore. Association for Computational Linguistics.

\bibitem[{Wolf et~al.(2020)Wolf, Debut, Sanh, Chaumond, Delangue, Moi, Cistac, Rault, Louf, Funtowicz, Davison, Shleifer, von Platen, Ma, Jernite, Plu, Xu, Le~Scao, Gugger, Drame, Lhoest, and Rush}]{wolf-etal-2020-transformers}
Thomas Wolf, Lysandre Debut, Victor Sanh, Julien Chaumond, Clement Delangue, Anthony Moi, Pierric Cistac, Tim Rault, Remi Louf, Morgan Funtowicz, Joe Davison, Sam Shleifer, Patrick von Platen, Clara Ma, Yacine Jernite, Julien Plu, Canwen Xu, Teven Le~Scao, Sylvain Gugger, Mariama Drame, Quentin Lhoest, and Alexander Rush. 2020.
\newblock \href {https://doi.org/10.18653/v1/2020.emnlp-demos.6} {Transformers: State-of-the-art natural language processing}.
\newblock In \emph{Proceedings of the 2020 Conference on Empirical Methods in Natural Language Processing: System Demonstrations}, pages 38--45, Online. Association for Computational Linguistics.

\bibitem[{Xie et~al.(2024)Xie, Wu, Jeon, He, Steck, Jha, Liang, Kallus, and McAuley}]{xie2024neighbor}
Zhouhang Xie, Junda Wu, Hyunsik Jeon, Zhankui He, Harald Steck, Rahul Jha, Dawen Liang, Nathan Kallus, and Julian McAuley. 2024.
\newblock Neighborhood-based collaborative filtering for conversational recommendation.
\newblock In \emph{RecSys}.

\bibitem[{Xu et~al.(2023)Xu, Wang, Mao, Lyu, She, and Zhang}]{xu2023knn}
Benfeng Xu, Quan Wang, Zhendong Mao, Yajuan Lyu, Qiaoqiao She, and Yongdong Zhang. 2023.
\newblock \href {https://openreview.net/forum?id=fe2S7736sNS} {\$k\${NN} prompting: Beyond-context learning with calibration-free nearest neighbor inference}.
\newblock In \emph{The Eleventh International Conference on Learning Representations}.

\bibitem[{Yao et~al.(2023)Yao, Zhao, Yu, Du, Shafran, Narasimhan, and Cao}]{yao2023react}
Shunyu Yao, Jeffrey Zhao, Dian Yu, Nan Du, Izhak Shafran, Karthik Narasimhan, and Yuan Cao. 2023.
\newblock {ReAct}: Synergizing reasoning and acting in language models.
\newblock In \emph{International Conference on Learning Representations (ICLR)}.

\bibitem[{Yu et~al.(2023)Yu, Chen, and Yu}]{yu-etal-2023-prompt}
Xiao Yu, Maximillian Chen, and Zhou Yu. 2023.
\newblock \href {https://doi.org/10.18653/v1/2023.emnlp-main.439} {Prompt-based {M}onte-{C}arlo tree search for goal-oriented dialogue policy planning}.
\newblock In \emph{Proceedings of the 2023 Conference on Empirical Methods in Natural Language Processing}, pages 7101--7125, Singapore. Association for Computational Linguistics.

\bibitem[{Zhang et~al.(2024)Zhang, Zhang, Rekabdar, Zhou, Wang, and Liu}]{Zhang2024DynamicAA}
XinHao Zhang, Jinghan Zhang, Banafsheh Rekabdar, Yuanchun Zhou, Pengfei Wang, and Kunpeng Liu. 2024.
\newblock \href {https://api.semanticscholar.org/CorpusID:270286174} {Dynamic and adaptive feature generation with llm}.
\newblock \emph{ArXiv}, abs/2406.03505.

\bibitem[{Zhao et~al.(2024)Zhao, Zhou, Li, Tang, Wang, Hou, Min, Zhang, Zhang, Dong, Du, Yang, Chen, Chen, Jiang, Ren, Li, Tang, Liu, Liu, Nie, and Wen}]{zhao2024surveylargelanguagemodels}
Wayne~Xin Zhao, Kun Zhou, Junyi Li, Tianyi Tang, Xiaolei Wang, Yupeng Hou, Yingqian Min, Beichen Zhang, Junjie Zhang, Zican Dong, Yifan Du, Chen Yang, Yushuo Chen, Zhipeng Chen, Jinhao Jiang, Ruiyang Ren, Yifan Li, Xinyu Tang, Zikang Liu, Peiyu Liu, Jian-Yun Nie, and Ji-Rong Wen. 2024.
\newblock \href {http://arxiv.org/abs/2303.18223} {A survey of large language models}.

\bibitem[{Zhong et~al.(2023)Zhong, Zhang, Li, Ahn, Klein, and Steinhardt}]{Zhong2023D5}
Ruiqi Zhong, Peter Zhang, Steve Li, Jinwoo Ahn, Dan Klein, and Jacob Steinhardt. 2023.
\newblock \href {http://papers.nips.cc/paper\_files/paper/2023/hash/7e810b2c75d69be186cadd2fe3febeab-Abstract-Conference.html} {Goal driven discovery of distributional differences via language descriptions}.
\newblock In \emph{Advances in Neural Information Processing Systems 36: Annual Conference on Neural Information Processing Systems 2023, NeurIPS 2023, New Orleans, LA, USA, December 10 - 16, 2023}.

\end{thebibliography}

\appendix

\section{Can \ourmethod~Faithfully Follow Users' Instruction?}
\label{sec:instruction_following}

The best way to evaluate whether \ourmethod~can adapt it's discovery based on users' goal is to observe the output of the framework on the same dataset with different user goals.
To this end, we conduct an additional experiment where we instruct \ourmethod~to categorize different aspect mentioned in a peer-review dataset, with the goal of discovery latent properties related to "clarity of writing" and "topics", respectively.
Specifically, we use the ACL and ICLR 2017 reviews that are associated with a "clarity" aspect from the PeerRead dataset~\cite{kang-etal-2018-dataset}, a dataset with peer-reviews collected from OpenReview.
This selection of dataset ensures the dataset contains valid variations in two directions: quality of writing and topic of contents.

To evaluate the success of our method in adapting to user goals, we prompt LLMs with a property generated by our framework, and let the LLM decide whether it is more relevant to the subject of writing style or content.
In this case, a higher accuracy indicates a better framework at following user instruction.
GPT-3.5 and GPT-4o has accuracy of recovering the goal from the generated topics at 83.9 and 90.1, respectively, indicating our framework can indeed adapts its discovery process based on user intruction.

\section{Discussion on Binarized Variant of~\ourmethod}
\label{sec:binarized_ours}

Since our method outputs continuous values for its latent space, one might wonder if this contributes to the superior performance of~\ourmethod, and if so, by how much.
Further, in some cases, users might want to understand our systems' decision on exactly whether a property is linked to a training data point (e.g. a document).
To this end, we included a binarized version of our method where we treat the top 10 percent of data-property links (sorted by the estimated compatibility score) as positive linkages with value 1, and else as negative linkages with value 0.
As shown in~\Cref{tab:embodied_navigation_metrics},~\Cref{tab:document_categorization_metrics}, and~\Cref{tab:movie_recommendation_metrics}, binarized variant of~\ourmethod's performance is slightly degraded, but nevertheless outperforms baselines methods consistent with the trends of vanilla~\ourmethod.

\section{Task Difficulties}
\label{sec:task_diffuculties}

The difficulty of each tasks we choose can be observed from the performance of a trivial "Majority" baseline that always picks the most popular outcome (a movie, a next action, or a document label).
As shown, all tasks requires models to learn the notion of data point similarities in latent space beyond trivially modeling popularity.

Another choice we made is converting the embodied navigation task on Alfworld to next action prediction.
We note this task, while eliminating other variations that affects a models' performance (e.g. an LLMs' compatibility to a particular prompt in navigation), is still a challenging task for LLMs.
In early stage of the experiments, we experimented with a ReAct~\cite{yao2023react} GPT-3.5-based agent on next action prediction.
This method cannot outperform~\ourmethod~on next-action prediction, even when enhanced with kNN-ICL~\cite{xu2023knn}, a recent state-of-the-art method for in-context example selection.

\section{Case Studies}
\label{sec:case_studies}

\begin{table*}[ht]
\centering
\begin{tabularx}{\textwidth}{lcX}
\toprule
\textbf{Dataset} & \textbf{Latent Factor} & \textbf{Correlated properties} \\
\midrule
\textbf{Insp.} & Dark Humor Movies & Absurdism, Absurdist elements, Absurdist humor, Blend of dark comedy, suspense, and drama, Blend of humor and deeper story elements, Clown Character, Clown Theme, Coen Brothers film, Comical Gore, Quirky or unique premise, Raunchy humor elements \\

\addlinespace %

\textbf{Alf.} & user needs to interact with Sofa & arby objects that matter for completing the task are the diningtable 1 and the sofa, relevant objects for completing the task are the coffeetable 1 and the sofa, sofa is present in the room and is relevant to the task completion. tasked with putting both remote controls in the sofa, user is currently located near the sofa'\\

\addlinespace

\textbf{Bills} & Civil Benefit & Cemetery Benefits, Central Intelligence Agency Retirement and Disability System, Child's Insurance Benefits under Social Security Church Organizations and Employee Benefits Civil Service Retirement System Cost-of-Living Allowances for Government Employees Death Benefits Elderly Financial Security\\
\bottomrule

\end{tabularx}
\caption{Discovered Latent Topics by~\ourmethod~in Three Different Datasets}
\label{tab:case_studies_ours}
\end{table*}

\begin{table*}[ht]
\centering
\begin{tabularx}{\textwidth}{lX}
\toprule
\textbf{Dataset} & Topics \\
\midrule
\textbf{Insp.} & User prefers critically acclaimed movies (Count: 3): the user tends to prefer movies that are critically acclaimed. \\

\addlinespace %

\textbf{Alf.} & Partially complete the task by addressing 1 item of 1 (Count: 37): The user has completed one step towards the task by cleaning the cup.\\

\addlinespace

\textbf{Bills} & Transportation (Count: 32): Mentions policies related to transportation benefits for employees.\\
\bottomrule

\end{tabularx}
\caption{Discovered Latent Topics by~TopicGPT~in Three Different Datasets}
\label{tab:case_studies_topicgpt}
\end{table*}

\begin{table*}[ht]
\centering
\begin{tabularx}{\textwidth}{lX}
\toprule
\textbf{Dataset} & Topics \\
\midrule
\textbf{Insp.} & user, system, that, documentary, about, it, you, of, any, the, and, movie, like, historical, interesting, good, to, action, in, would \\

\addlinespace %

\textbf{Alf.} & ottoman, laptop, bowl, plate, newspaper, 25, sofa, coffeetable, diningtable, vase, 18, keychain, tissuebox, armchair, pencil, remotecontrol, drawer, inon, book, two\\

\addlinespace

\textbf{Bills} & 'phosphate', 'phosphor', 'lanthanum', 'harmonized', 'tariff', 'suspension', 'schedule', 'yttrium', 'duty', 'oxide', 'coprecipitates', 'cerium', 'activated', 'extend', 'temporary', 'europium', 'terbium', 'magnesium', 'united', 'temporarily'\\
\bottomrule

\end{tabularx}
\caption{Discovered Latent Topics by~BertTopic~in Three Different Datasets}
\label{tab:case_studies_BertTopic}
\end{table*}

\begin{table*}[ht]
\centering
\begin{tabularx}{\textwidth}{lX}
\toprule
\textbf{Dataset} & Topics \\
\midrule
\textbf{Insp.} & system, user, movie, good, nice, think, movies, yes, new, action, watching, trailer, see, know, called, screenplay, talking, enjoy, love, stanley \\

\addlinespace %

\textbf{Alf.} & countertop, cup, task, put, matter, completing, nearby, user, objects, given, cool, still, per, needs, cooled, accomplished, successfully, process, fridge, picking\\

\addlinespace

\textbf{Bills} & patient, making, decision, providers, expert, understanding, item, specify, require, aids, implemented, secretary, timely, consultation, shared, establish, act, considering, steps, service\\
\bottomrule

\end{tabularx}
\caption{Discovered Latent Topics by~LDA~in Three Different Datasets}
\label{tab:case_studies_LDA}
\end{table*}

Example non-cherry-picked latent topic discovered by~\ourmethod~and baselines are as shown in \Cref{tab:case_studies_ours},~\Cref{tab:case_studies_topicgpt},~\Cref{tab:case_studies_BertTopic},~\Cref{tab:case_studies_LDA}. 
In each of the evaluation scenarios, our framework can identify a set of correlated properties following a coherent theme, and produce more meaningful results than the description of a topic as produced by TopicGPT.
Importantly, these properties are grouped together in a task-oriented manner.
For example, the the case of Alfworld, our framework can identify various concepts related to ``user needs to interact with sofa", a task-relevant concept. 
This is in contrast to outputs from BertTopic and LDA, where the keywords do not always follow the users' goal and contain irrelevant words.

\section{Open Source Models}
\label{sec:disuccsion_open_source_models}

In~\citeauthor{pham-etal-2024-topicgpt}'s work, it was observed that opensource LLM, such as Mistral-7B~\cite{jiang2023mistral7b}, does not have strong enough reasoning ability to correctly organize the generated concepts, and as a result causes TopicGPT to fail to produce coherent results.
We hypothesize that this is due to the complexity of model instructions in TopicGPT.
To this end, we conduct an additional experiment on the Inspired dataset to see if our method can make open-source models have good performance.
Using Mistral-7B~\cite{jiang2023mistral7b}, our method continues to have stable performance across datasets.
This shows that our method indeed can mitigate the reliance on a stronger LLM than prior works.
In contrast, we also run another variant of TopicGPT, TopicGPT-4 with GPT-4 on movie recommendation, but find it cannot help the pure-LLM based framework to handle noisy dialogues.
The $\mathrm{Hit}@1,5,20$ are 0.48, 1.40, and 2.40 respectively, which is comparable to other variants of TopicGPT and less performant than~\ourmethod.

\section{Comparison Between LLMs on~\ourmethod}
In our experiments, the LLMs (GPT-3.5, GPT-4o, Mistral) show comparable performance, and there is no single method that's the best performing across all scenarios.
We hypothesize that this is due to our \texttt{property proposal} step is not dependent on an LLMs' reasoning ability, and thus, these recent LLMs can all provide viable results for our method (in contrast to prior work such as TopticGPT).

\section{Discussion on Prompts Used}
\label{sec:discussion_on_prompts}

We provide the prompts we used on each dataset for~\ourmethod~(\Cref{tab:inspired-prompt},~\Cref{tab:alfworld-prompt}, and ~\Cref{tab:bills-prompt}) and TopicGPT (\Cref{tab:topicgpt-inspired-prompt},~\Cref{tab:topicgpt-alfworld-prompt}, and ~\Cref{tab:topicgpt-bills-prompt}).
As shown, since~\ourmethod~only requires the LLM to propose interesting properties, the prompt for~\ourmethod~is shorter, contains less instructions for different situations, and relies less on in-context examples.
We note that this is a core advantage of our method, in that our method are less demanding on a strong LLM that can faithfully follow various instructions.

To evaluate the stability of our prompt, we conduct an additional experiments on the Inspired dataset, where we use GPT-3.5 to rephrase all the prompts we use for~\ourmethod-3.5.
The $\mathrm{Hit}@k \in {1, 5, 20}$ is 3.36, 13.46, and 22.11, respectively.
This shows that our framework is robust to variation of prompts.

\section{Implementation Details}
\label{sec:implementation_details}

\paragraph{Dataset Statistics}

The datapoints used for training and evaluation is as described in~\Cref{sec:movie_recommendation},~\Cref{sec:embodied_navigation}, and ~\Cref{sec:document_categorization}

\paragraph{Model Parameters and Computing Resources}

In this section we list the parameters of our models.
GPT-3.5 and GPT-4o are propriety models whose parameter is unknown. 
The sentence embedding model we use across all experiments is a small distilled embedding model with state-of-the-art performance\footnote{sentence-transformers/all-MiniLM-L6-v2 on Huggingface Transformers library~\cite{wolf-etal-2020-transformers}}, containing 22.7 million paramters.
For Linear Corex, we use the implementation open-sourced by~\citeauthor{Steeg19linearcorex, Steeg2017FastSL}, available at~\url{https://github.com/hrayrhar/T-CorEx}.
We note that Linear Corex is a linear model whose parameter weights is equal to the number (types) of properties generated by LLMs on each dataset.
There are 8569 unique concepts generated on Inspired, 41020 unique property generated on Alfworld, and 7580 unique concepts generated on bills dataset.
Following prior works in crowd-sourcing labels for a dataset~\cite{Bragg2013CrowdsourcingMC}, we note that there can be additional methods to speed-up the property generation process, e.g. by estimating the number of novel properties that won't be generated in the future; we leave this to future works.

Experiments are conducted on a server with Nvidia RTX A6000 GPUs with 48GB memory each.

\paragraph{Hyperparameters}

Unless specifically specified, we use default hyper-parameters in the above-discussed code-bases and libraries we use in this work).

\section{Further Related Work}
\label{sec:further_related_work}

Aside from core related works discussed in the main content sections, our work is also in-line with goal-oriented clustering with LLMs~\cite{wang-etal-2023-goal}, LLM for label taxonomy creation~\cite{wan2024tntlm}, and LLM for feature engineering~\cite{Zhang2024DynamicAA, oikarinen2023labelfree}.
However, these works focus on other setting than latent factor discovery (or closely related area such as topic modeling).
Our work also shares insights with concept bottleneck models~\cite{koh2020conceptbottleneckmodels}, label-free concept bottleneck models~\cite{oikarinen2023labelfreeconceptbottleneckmodels}, and related concept-based explainable machine learning models, but focuses on \textit{discovering} informative concepts rather than leveraging concepts for explainability.

\section{Additional Explanation for Our Framework}
\label{sec:our_framework_detailed}

\begin{figure*}[htb]
\centering
\includegraphics[scale=0.65]{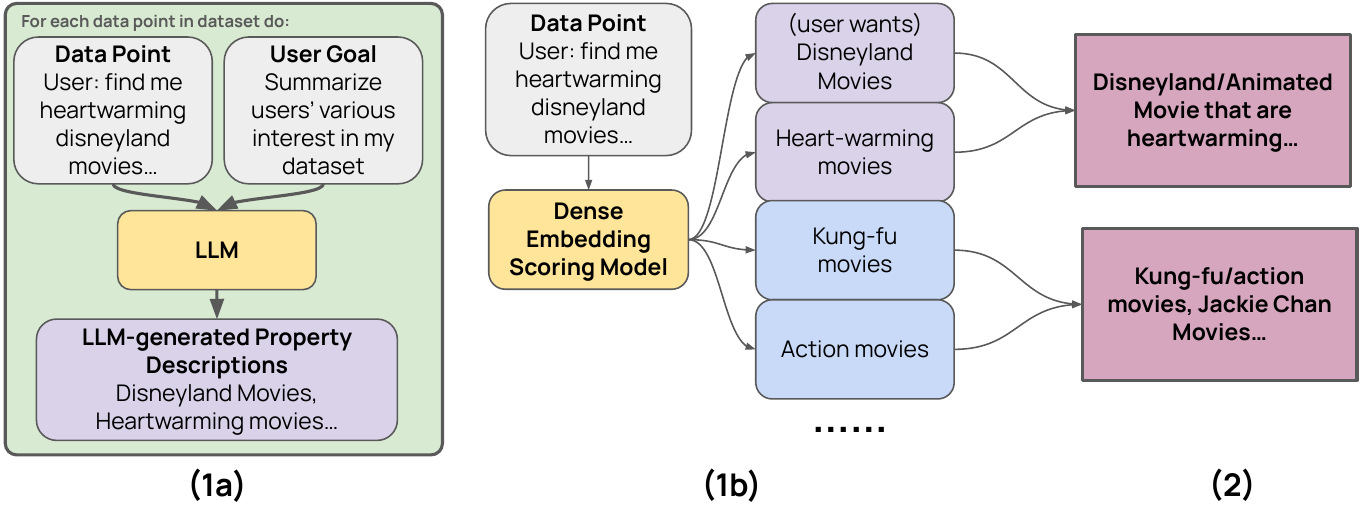}
\caption{The proposed framework with concrete examples. See~\Cref{sec:our_framework_detailed} for discussion.
}
\label{fig:our_framework_detailed}
\vspace{-2mm}
\end{figure*}

In this section, we provide additional explanation for our framework using a more detailed figure (\Cref{fig:our_framework_detailed}) to~\Cref{fig:our_framework}. For each data point (document) in the copus, we first prompt LLM with the users’ goal to generate a few key properties that describes the characteristics of the document (1a). After this, we use a dense embedding model to estimate the compatibility score between each document and \textit{all} possible properties that ever generated (even for properties generated from other document, such as kung-fu movies). Finally, highly-correlated properties (based on estimated scores in 1b) will be grouped, representing higher-level concepts.

\section{Human Evaluation Details}
\label{sec:human_evaluation_details}

\begin{figure*}[tb]
\centering
\includegraphics[scale=0.65]{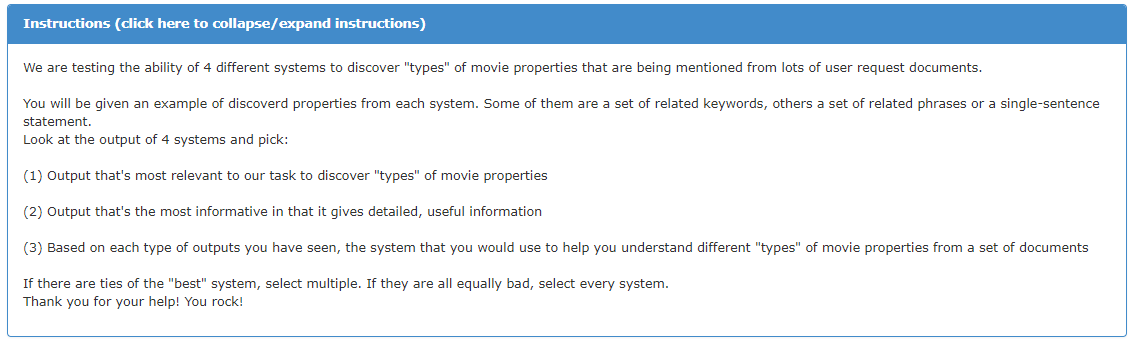}
\caption{The instruction for our main human evaluation, results as shown in~\Cref{tab:human_relevance_informativeness}
}
\label{fig:turker_instruction_main_exp}
\end{figure*}

\begin{figure*}[tb]
\centering
\includegraphics[scale=0.65]{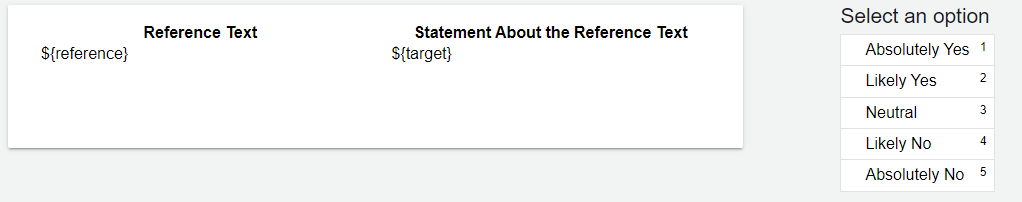}
\caption{The instruction for our human evaluation to verify that LLMs are better concept generators than assigners.
}
\label{fig:turker_instruction_gen_ai_paradox}
\end{figure*}

\begin{figure*}[tb]
\centering
\includegraphics[scale=0.65]{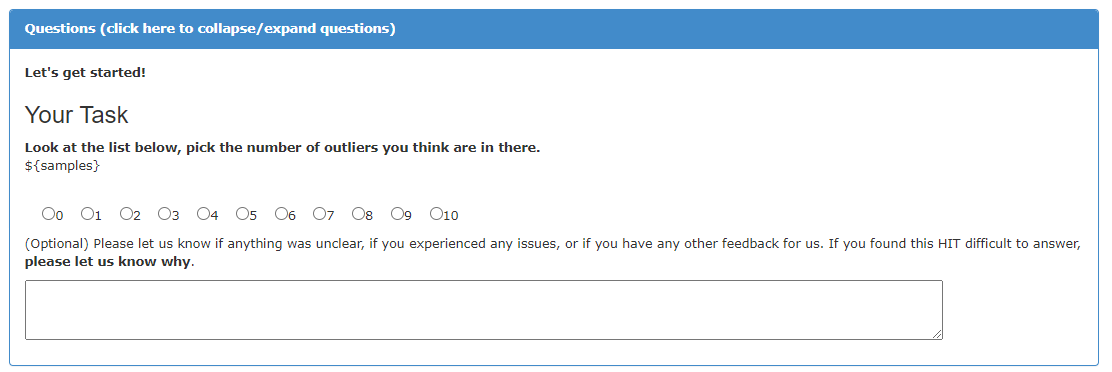}
\caption{The instruction for our human evaluation to check the number of outliers among each learned latent pattern. Results are as shown in~\Cref{tab:observation_grouping_qual}
}
\label{fig:turker_instruction_outlier_counts}
\end{figure*}

We perform our human evaluation on Amazon Mechanical Turk (\url{https://www.mturk.com/}). 
We request for workers with a life-time approval rate of 95\% or beyond from United States.
Instructions to the Turk crowd workers is as shown in~\Cref{fig:turker_instruction_main_exp},~\Cref{fig:turker_instruction_gen_ai_paradox}, and~\Cref{fig:turker_instruction_outlier_counts}.
The majority of single turker tasks (~\Cref{fig:turker_instruction_main_exp},~\Cref{fig:turker_instruction_gen_ai_paradox}) are awarded 1 dollar per evaluation, which takes a minte to two.
Outlier detection ask (~\Cref{fig:turker_instruction_outlier_counts}) takes less than 1 minute so we pay 0.5 dollar per task.
These results yields approximately doubles the baseline wage in most states (e.g. Arizona, Colorado).

\begin{table*}[t!]
\small
\center
\begin{tabular}{l}
\toprule
\bf Prompt for~\ourmethod{}~on Inspired  \\
\midrule
\begin{minipage}[t]{2.0\columnwidth}
\textbf{Prompt to generate initial property}: help me analyze the following dialogue between the user and a movie recommender assistant. We are particularly interested in factors that affects what movie should the assistant discuss/recommend in the next response. Pay special attention to the task the current topic and the users' expressed interest in movie. \newline\newline

Here is the interaction log:\newline
<request> \newline\newline

Generate a one sentence description of the dialogue's current state w.r.t. what type of movie to recommend next. what's the users' preference? are any properties of the next movie to discuss known to us?\newline
---\newline
\textbf{Prompt to format initial property}: Now, given your inferred current dialogue situation, propose a numbered list of property keywords that the next movie being discussed likely satisfy. E.g., "Romantic Genre", "Comedy Genre", "Features actor X", "Superhero movie", "Dark humor elements", etc. Your numbered list of properties:\newline
\end{minipage} \\ 

\bottomrule
\end{tabular}
\caption{Prompts we use on Inspired.
}
\label{tab:inspired-prompt}
\end{table*}

\begin{table*}[t!]
\small
\center
\begin{tabular}{l}
\toprule
\bf Prompt for~\ourmethod{}~on Alfworld  \\
\midrule
\begin{minipage}[t]{2.0\columnwidth}
\textbf{Prompt to generate initial property}: help me analyze the following action log ("input)" of a user. We are particularly interested in factors that affects what the user would do next. Pay special attention to the task the user is given and the users' newest state in the interaction log.\newline\newline 

Here is the interaction log:\newline
<InteractionLog> \newline\newline

Generate a one sentence description of the users' current process w.r.t. completing the given task - what still needs to be accomplished? what's already accomplished? Any nearby objects/utensils that matters for completing the task? what's the user's given task?\newline
---\newline
\textbf{Prompt to format initial property}: Now, given your inferred current user situation, propose a numbered list of property keywords that describes the users' current status w.r.t. task completion, e.g. "already cleaned an item", "a microwave is at current location", "the user is tasked with cleaning an item", "the user needs to find a pot", etc. Your numbered list of properties:\newline
\end{minipage} \\ 

\bottomrule
\end{tabular}
\caption{Prompts we use on Alfworld.
}
\label{tab:alfworld-prompt}
\end{table*}

\begin{table*}[t!]
\small
\center
\begin{tabular}{l}
\toprule
\bf Prompt for~\ourmethod{}~on Bills  \\
\midrule
\begin{minipage}[t]{2.0\columnwidth}
\textbf{Prompt to generate initial property}: help me analyze the following bill summary from U.S. congresses. We are particularly interested in factors that governs the topic this document addresses, e.g. trades, foreign trades, agriculture, etc.\newline\newline

Here is the document:\newline
<document> \newline\newline

Generate a one sentence description of the key topics/directions addressed by this document.\newline
---\newline
\textbf{Prompt to format initial property}: Now, given your inferred current user situation, propose a numbered list of property keywords that describes the users' current status w.r.t. task completion, e.g. "already cleaned an item", "a microwave is at current location", "the user is tasked with cleaning an item", "the user needs to find a pot", etc. Your numbered list of properties:\newline
\end{minipage} \\ 

\bottomrule
\end{tabular}
\caption{Prompts we use on Inspired (GPT-rewritten).
}
\label{tab:bills-prompt}
\end{table*}

\begin{table*}[t!]
\small
\center
\begin{tabular}{l}
\toprule
\bf Prompt for~\ourmethod{}~on Bills  \\
\midrule
\begin{minipage}[t]{2.0\columnwidth}
\textbf{Prompt to generate initial property}: Assist me in analyzing the following dialogue between a user and a movie recommendation assistant. Specifically, identify the key factors that influence which movie the assistant should recommend next. Focus on the current task, the conversation's topic, and the user's expressed preferences.\newline\newline

Below is the interaction log: <request>\newline

Generate a brief one-sentence summary of the dialogue’s current state, particularly regarding the type of movie to recommend next. What are the user's preferences? Do we know any characteristics of the next movie to discuss?\newline
---\newline
\textbf{Prompt to format initial property}: Based on the current state of the dialogue you've inferred, create a numbered list of property keywords that the next movie being discussed is likely to match. For example: "Romantic Genre," "Comedy Genre," "Features actor X," "Superhero movie," "Dark humor elements," etc. Your list of properties:\newline
\end{minipage} \\ 

\bottomrule
\end{tabular}
\caption{Prompts we use on Inspired, rewritten from the original prompt by GPT-3.5, to analyze the prompt-stability of~\ourmethod.
}
\label{tab:inspired-rewritten-prompt}
\end{table*}

\begin{table*}[t!]
\small
\center
\begin{tabular}{l}
\toprule
\bf Prompt for TopicGPT on Inspired  \\
\midrule
\begin{minipage}[t]{2.0\columnwidth}
You will receive a document that is a conversation log between user and system. Your task is to identify generalizable traits (topics) that can act as top-level topics in the hierarchy. If any relevant topics are missing from the provided set, please add them. Otherwise, output the existing top-level topics as identified in the document.\newline
A topic in this case is a characteristic of the current dialogue history that determines what movie we should recommend next to the user.\newline\newline

[Top-level topics]\newline
{Topics}\newline\newline

[Instructions]\newline
Step 1: Determine topics mentioned in the document. \newline
- The topic labels must be as GENERALIZABLE as possible. They must not be document-specific.\newline
- The topics must reflect a SINGLE topic instead of a combination of topics.\newline
- The new topics must have a level number, a short general label, and a topic description. \newline
- The topics must be broad enough to accommodate future subtopics. \newline
Step 2: Perform ONE of the following operations: \newline
1. If there are already duplicates or relevant topics in the hierarchy, output those topics and stop here. \newline
2. If the document contains no topic, return "None". \newline
3. Otherwise, add your topic as a top-level topic. Stop here and output the added topic(s). DO NOT add any additional levels.\newline\newline

[Examples]\newline
Example 1: Adding "[1] User expressed interest in action movies"\newline
Document: \newline
System: what movies do you like? User: I like action movies! System:\newline\newline

Your response: \newline
[1] User expressed interest in action movies: the user expressed in some way that he/she enjoys action movie.\newline\newline

Example 2: Duplicate "[1] User expressed interest in action movies", returning the existing topic"\newline
Document: \newline
System: Hi there, can you tell me what movies you typically watch? User: Well, i watch a bunch of action movies after work usually. System:\newline\newline

Your response: \newline
[1] User expressed interest in action movies", returning the existing topic\newline\newline

[Document]\newline
{Document}\newline\newline

Focus on topics that are relevant to what the system should recommend text. Please ONLY return the relevant or modified topics at the top level in the hierarchy.\newline
[Your response]\newline
\end{minipage} \\ 

\bottomrule
\end{tabular}
\caption{Prompts TopicGPT~\cite{pham-etal-2024-topicgpt} use on Inspired.
}
\label{tab:topicgpt-inspired-prompt}
\end{table*}

\begin{table*}[t!]
\small
\center
\begin{tabular}{l}
\toprule
\bf Prompt for TopicGPT on Alfworld  \\
\midrule
\begin{minipage}[t]{2.0\columnwidth}
You will receive a document that is an interaction log between user and an environment. Your task is to identify generalizable traits (topics) that can act as top-level topics in the hierarchy. If any relevant topics are missing from the provided set, please add them. Otherwise, output the existing top-level topics as identified in the document.\newline
A topic in this case is a characteristic of the users' current state w.r.t. the given task that helps us reason about what should the user do next.\newline\newline

[Top-level topics]\newline
{Topics}\newline\newline

[Instructions]\newline
Step 1: Determine topics mentioned in the document. 
- The topic labels must be as GENERALIZABLE as possible. They must not be document-specific.\newline
- The topics must reflect a SINGLE topic instead of a combination of topics.\newline
- The new topics must have a level number, a short general label, and a topic description. \newline
- The topics must be broad enough to accommodate future subtopics. \newline
Step 2: Perform ONE of the following operations: \newline
1. If there are already duplicates or relevant topics in the hierarchy, output those topics and stop here. \newline
2. If the document contains no topic, return "None". \newline
3. Otherwise, add your topic as a top-level topic. Stop here and output the added topic(s). DO NOT add any additional levels.\newline\newline

[Examples]\newline
Example 1: Adding "[1] The user just started the task: the user has not take any action yet."\newline
Document: \newline
{"text": "[{'state': '-= Welcome to TextWorld, ALFRED! =-You are in the middle of a room. Looking quickly around you, you see a bed 1, a desk 1, a drawer 8, a drawer 7, a drawer 6, a drawer 5, a drawer 4, a drawer 3, a drawer 2, a drawer 1, a dresser 1, a garbagecan 1, a shelf 5, a shelf 4, a shelf 3, a shelf 2, and a shelf 1.Your task is to: find two bowl and put them in desk.'}]"}\newline\newline

Your response: \newline
[1] The user just started the task: the user has not take any action yet.\newline\newline

Example 2: Adding "[1] Partially complete the task by addressing 1 item of 2: The user has dopped some of the required item by the task to the target location"
Document:\newline
[{'state': '-= Welcome to TextWorld, ALFRED! =-You are in the middle of a room. Looking quickly around you, you see a bed 1, a desk 1, a drawer 8, a drawer 7, a drawer 6, a drawer 5, a drawer 4, a drawer 3, a drawer 2, a drawer 1, a dresser 1, a garbagecan 1, a shelf 5, a shelf 4, a shelf 3, a shelf 2, and a shelf 1.Your task is to: find two bowl and put them in desk.'}, ('action', 'go to shelf 3'), {'state': 'You arrive at loc 21. On the shelf 3, you see a bowl 1, and a creditcard 1.'}, ('action', 'take bowl 1 from shelf 3'), {'state': 'You pick up the bowl 1 from the shelf 3.'}, ('action', 'go to desk 1'), {'state': 'You arrive at loc 18. On the desk 1, you see a laptop 1, and a pen 2.'}, ('action', 'put bowl 1 in/on desk 1'), {'state': 'You put the bowl 1 in/on the desk 1.'}]\newline\newline

Your response: \newline
[1] Partially complete the task by addressing 1 item of 2: The user has dopped some of the required item by the task to the target location\newline

Now, look at the following document, and generate your description about the users' status w.r.t. the task.\newline

[Current Document]\newline
{Document}\newline

Now, give me description about the users' current state w.r.t. completing the given task in the document. Please ONLY return the relevant or modified topics at the top level in the hierarchy.\newline
Remember, your response should start with the level of the topic, e.g. [1]: <short description>:<details>\newline
[Your response]\newline
\end{minipage} \\ 

\bottomrule
\end{tabular}
\caption{Prompts TopicGPT~\cite{pham-etal-2024-topicgpt} use on Alfworld.
}
\label{tab:topicgpt-alfworld-prompt}
\end{table*}

\begin{table*}[t!]
\small
\center
\begin{tabular}{l}
\toprule
\bf Prompt for TopicGPT on Bills  \\
\midrule
\begin{minipage}[t]{2.0\columnwidth}
You will receive a document and a set of top-level topics from a topic hierarchy. Your task is to identify generalizable topics within the document that can act as top-level topics in the hierarchy. If any relevant topics are missing from the provided set, please add them. Otherwise, output the existing top-level topics as identified in the document.\newline\newline

[Top-level topics]\newline
{Topics}\newline\newline

[Examples]\newline
Example 1: Adding "[1] <topic-label>"\newline
Document: \newline
<doc-example-1>\newline\newline

Your response: \newline
[1] <topic-label>: <topic-desc>\newline\newline

Example 2: Duplicate "[1] <topic-label>", returning the existing topic\newline
Document: \newline
<doc-example-2>\newline\newline

Your response: \newline
[1] <topic-label>: <topic-desc>\newline\newline

[Instructions]\newline
Step 1: Determine topics mentioned in the document. \newline
- The topic labels must be as GENERALIZABLE as possible. They must not be document-specific.\newline
- The topics must reflect a SINGLE topic instead of a combination of topics.\newline
- The new topics must have a level number, a short general label, and a topic description. \newline
- The topics must be broad enough to accommodate future subtopics. \newline
Step 2: Perform ONE of the following operations: \newline
1. If there are already duplicates or relevant topics in the hierarchy, output those topics and stop here. \newline
2. If the document contains no topic, return "None". \newline
3. Otherwise, add your topic as a top-level topic. Stop here and output the added topic(s). DO NOT add any additional levels.\newline\newline

[Document]\newline
{Document}\newline\newline

Please ONLY return the relevant or modified topics at the top level in the hierarchy.\newline
[Your response]\newline
\end{minipage} \\ 

\bottomrule
\end{tabular}
\caption{Prompts TopicGPT~\cite{pham-etal-2024-topicgpt} use on Bills.
}
\label{tab:topicgpt-bills-prompt}
\end{table*}

\begin{figure}[tb]
\centering
\includegraphics[scale=0.6]{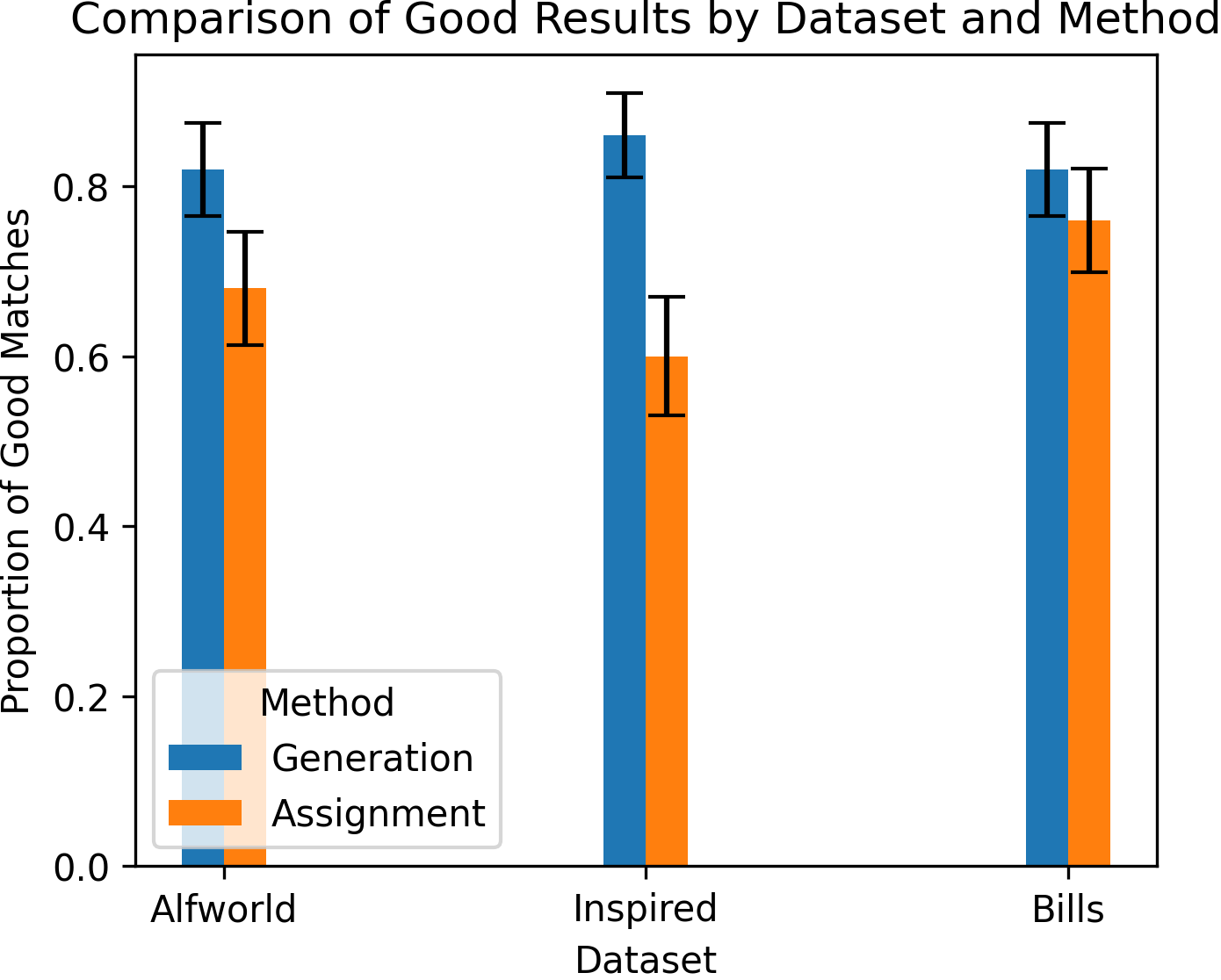}
\caption{LLMs are better property proposers than generators. This figure shows results on GPT-4o.
}
\label{fig:llm_are_better_generator_than_assigner_4o}
\vskip -5mm
\end{figure}

\section{On the efficiency and cost of~\ourmethod}
\label{sec:discussion_on_efficiency}

Finally, we demonstrate that~\ourmethod~is an efficient method in that all components in our framework are parallelizable. 
This is in contrast to~\citeauthor{pham-etal-2024-topicgpt}, where the topic/property proposal phase is non-parallelizable.
For example, on the Alfworld dataset, we show that the topic-proposal steps of~\ourmethod~finish in 60 minutes (including training of the embedding model), resulting in 50 times speed-ups compared to TopicGPT.

Since our method scales well to weaker, open-source models, one can run over framework without incurring API costs.
Still, non of our experiment runs on GPT-3.5 and GPT-4o costs more than 5 dollars, as of October 2024.

\section{Detail for Linear Corex}
\label{sec:detail_for_steeg}

Let $\nu_{C_i \mid Z}$ be the conditional mean of $C_i$ given $Z$ under the factorization $p(c_i|z) = p(c_i)/p(z) \Pi_j p(z_j|c_i)$, which is implied by the modular $TC(Z|C_i)=0$ constraint, the specific loss function that Linear Corex~\cite{Steeg19linearcorex} optimizes for is: $$\sum_{i=1}^p Q_i  \frac{1}{2} \log E\left[(C_i - \nu_{C_i \mid Z})^2\right] + \sum_{j=1}^m \frac{1}{2} \log E\left[Z_j^2\right].$$

\section{Potential Risks and Ethical Concerns}

We note that LLM are known to suffer from hallucinations in its generated content~\cite{zhao2024surveylargelanguagemodels}.
To this end, we advise practitioners to carefully verify the generated content from our framework before deploying it in critical decision-making scenarios.

\section{AI Assistant Usage Statement}

Writings in this work benefited from Grammarly (\url{https://app.grammarly.com/}) for grammar suggestions.

\section{Potential Risks and Ethical Concerns}

We note that LLM are known to suffer from hallucinations in its generated content~\cite{zhao2024surveylargelanguagemodels}.
To this end, we advise practitioners to carefully verify the generated content from our framework before deploying it in critical decision-making scenarios.

\end{document}